\documentclass[letterpaper, 10 pt, journal, twoside]{IEEEtran}
\IEEEoverridecommandlockouts                              % This command is only needed if 
\usepackage{graphicx, tipa}
\usepackage{amsmath,amssymb,amsfonts}
\usepackage[caption=false]{subfig}
\usepackage[ruled,linesnumbered, noend]{algorithm2e}
\usepackage{dirtytalk}
\usepackage{hyperref}
\usepackage{gensymb}
\usepackage{tabularx}
\usepackage{multirow}
\usepackage[short]{optidef}
\usepackage{booktabs}
\usepackage{siunitx}
\usepackage{titlesec}
\newcounter{cases}
\newcounter{subcases}[cases]

\makeatletter
\newcommand{\removelatexerror}{\let\@latex@error\@gobble}
\makeatother
\SetKwComment{Comment}{$\triangleright$\ }{}

\newcommand\Tstrut{\rule{0pt}{2.0ex}}         % = `top' strut
\title{Intent Prediction-Driven Model Predictive Control for UAV Planning and Navigation in Dynamic Environments} 

\author{Zhefan Xu\footnotemark*, Hanyu Jin\footnotemark*, Xinming Han, Haoyu Shen, and Kenji Shimada %<-this % stops a space
\thanks{Manuscript received: January, 6, 2025; Revised March, 10, 2025; Accepted March, 13, 2025.}
\thanks{This paper was recommended for publication by Editor Cesar Cadena upon evaluation of the Associate Editor and Reviewers' comments.
This paper is based on results obtained from a project of Programs for Bridging the gap between R\&D and the IDeal society (society 5.0) and Generating Economic and social value (BRIDGE)/Practical Global Research in the AI×Robotics Services, implemented by the Cabinet Office, Government of Japan.}
\thanks{Zhefan Xu, Hanyu Jin,  Xinming Han, Haoyu Shen, and Kenji Shimada are with the Department of Mechanical Engineering, Carnegie Mellon University, 5000 Forbes Ave, Pittsburgh, PA, 15213, USA.
        {\tt\footnotesize zhefanx@andrew.cmu.edu}}%
\thanks{*The authors contributed equally.}
\thanks{Digital Object Identifier (DOI): see top of this page.}
}
\begin{document}
\markboth{IEEE Robotics and Automation Letters. Preprint Version. Accepted March, 2025}
{Xu \MakeLowercase{\textit{et al.}}: Intent Prediction-Driven Model Predictive Control for UAV Planning and Navigation in Dynamic Environments} 
% \title{\large Bare Demo of IEEEtran.cls \\ for IEEE Conferences}

\maketitle

\noindent \begin{abstract}
Aerial robots can enhance construction site productivity by autonomously handling inspection and mapping tasks. However, 
ensuring safe navigation near human workers remains challenging. While navigation in static environments has been well studied, navigating dynamic environments remains open due to challenges in perception and planning. Payload limitations restrict the robots to using cameras with limited fields of view, resulting in unreliable perception and tracking during collision avoidance. Moreover, the rapidly changing conditions of dynamic environments can quickly make the generated optimal trajectory outdated.To address these challenges, this paper presents a comprehensive navigation framework that integrates perception, intent prediction, and planning. Our perception module detects and tracks dynamic obstacles efficiently and handles tracking loss and occlusion during collision avoidance. The proposed intent prediction module employs a Markov Decision Process (MDP) to forecast potential actions of dynamic obstacles with the possible future trajectories. Finally, a novel intent-based planning algorithm, leveraging model predictive control (MPC), is applied to generate navigation trajectories. Simulation and physical experiments\footnote{Experiment video link: \url{https://youtu.be/4xsEeMB9WPY}} demonstrate that our method improves the safety of navigation by achieving the fewest collisions compared to benchmarks.
\end{abstract}

\section{Introduction}
\IEEEPARstart{L}{ightweight} unmanned aerial vehicles (UAVs) have shown great potential for inspection and mapping in indoor construction sites \cite{construction-app1}\cite{construction-app2}\cite{construction-app3}. Unlike open outdoor areas, indoor sites contain moving workers, machinery, and other obstacles in close proximity, making operational safety a critical concern.  Consequently, a navigation framework with improved safety in these dynamic environments is essential.

Despite the success of prior research \cite{fast_planner}\cite{ego_planner} in complex static environments, safe navigation in dynamic settings remains challenging for two main reasons. First, the low image quality of onboard cameras in lightweight UAVs makes accurate detection and tracking of dynamic obstacles difficult. Although our previous work \cite{detector} utilizes an ensemble detection method to improve accuracy and efficiency, the camera's limited field of view (FOV) can still cause tracking failures during collision avoidance, increasing the risk of severe collisions with undetected dynamic obstacles. Second, the uncertain motion of dynamic obstacles can quickly make an optimized trajectory ineffective. While many works \cite{vision-mpc}\cite{zju_collision_avoidance}\cite{ViGO}\cite{chen2023flying} incorporate obstacle trajectory prediction into planning, they generally assume a single future mode (i.e., one trajectory) per obstacle, typically using a linear motion model. While these approaches can improve dynamic obstacle avoidance, effectiveness drops significantly if the actual motion of the obstacles differs from the assumed model.

\begin{figure}[t] 
    \vspace{0.1cm}
    \centering
    \includegraphics[scale=0.658]{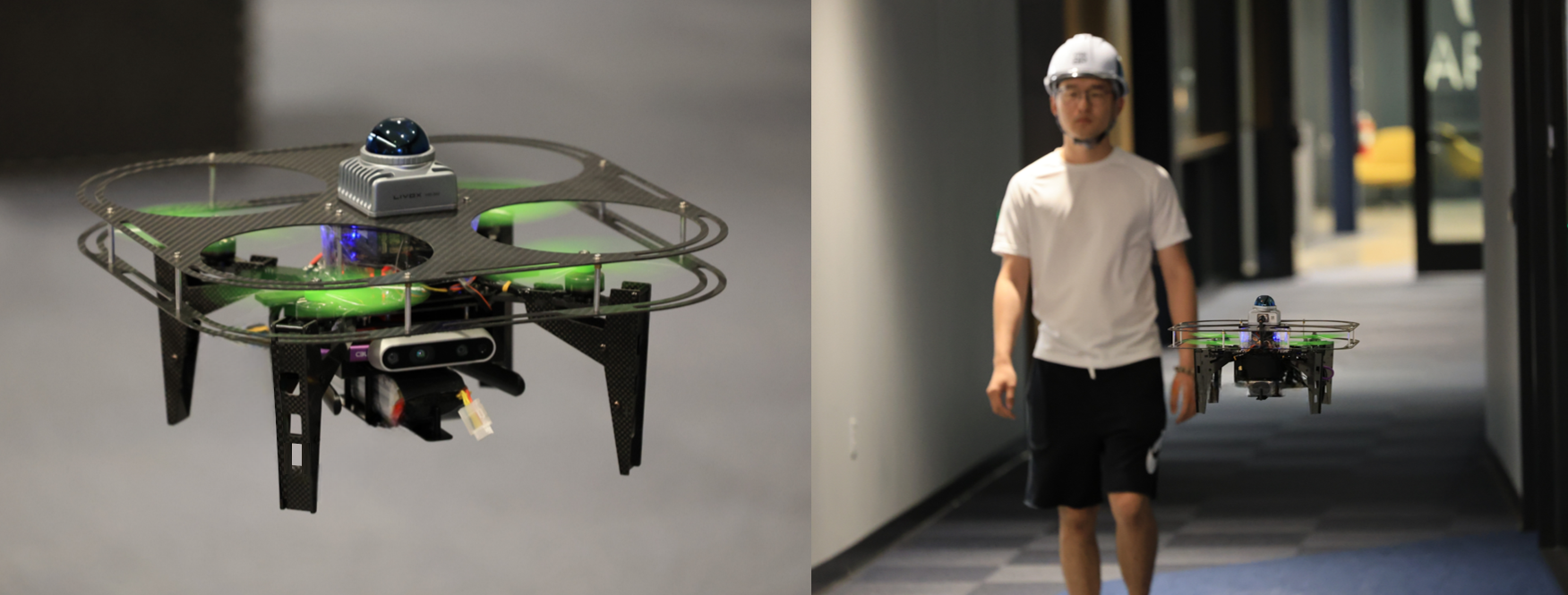}
    \caption{The UAV navigating a dynamic environment using the proposed algorithm. The left figure shows our customized UAV equipped with onboard sensors and the right figure illustrates the UAV avoiding a pedestrian. }
    \label{intro-figure}
\end{figure}

To address these challenges, this paper presents an intent prediction-driven navigation framework based on a model predictive control scheme. A perception module for dynamic obstacle detection and tracking is introduced to mitigate the lost-tracking issue during collision avoidance maneuvers. Crucially, to enable the planner to consider all potential future movements of dynamic obstacles, this work utilized dynamic obstacle intents, representing high-level behaviors such as turning, stopping, or moving straight, modeled as a probability distribution. By modeling each obstacle’s intent as a probability distribution over these possible actions, we capture the uncertainties more effectively. Instead of predicting a single trajectory per obstacle, the prediction module generates all possible trajectories reflecting the various intents. The proposed planning algorithm optimizes multiple trajectories according to future states using model predictive control and applies a scoring function to select the best trajectory. Fig. \ref{intro-figure} demonstrates our UAV navigating a dynamic environment. The contributions of this work are: 
\begin{itemize}
    \item \textbf{Intent Prediction-Driven Navigation Framework:} We introduce an intent prediction-driven navigation framework comprising three key components: (1) a perception module detecting both static and dynamic obstacles with lost-tracking awareness, (2) a prediction module forecasting dynamic obstacle intents and trajectories, and (3) a planning and control module using model predictive control to generate trajectories.  The  framework is made available as an open-source package on GitHub \footnote{Software available at: \url{https://github.com/Zhefan-Xu/Intent-MPC}}.
    \item \textbf{Intent and Trajectory Prediction:} This work adopts a Markov Decision Process (MDP) for dynamic obstacle intent prediction to achieve high-level obstacle action forecasting and future trajectory prediction. Unlike the previous MDP-based approach \cite{mdp_prediction}, which relies on long-term data to model task-level intent and is therefore less suitable for real-time navigation, our method enables short-term prediction of motion-level intent.
    \item \textbf{Physical Flight Experiments:} We validate the effectiveness of our proposed framework through extensive physical flight experiments in dynamic environments. The results demonstrate that our method enables the robot to navigate safely in various conditions.
\end{itemize}

\section{Related Work}
\textbf{Reactive methods:} These methods treat dynamic obstacles as static at each time step, adapting static planning methods to dynamic environments through high-frequency replanning. In \cite{fast_planner}\cite{ego_planner}, occupancy-based maps are adopted for environment representation, which enables safe trajectory generation. Ren et al. \cite{wholebody} achieve aggressive maneuvering using whole-body motion planning. However, these static-focused methods, lacking dynamic obstacle perception and planning, can fail in dynamic scenarios. To address the future uncertainties of dynamic obstacles, Guo et al. \cite{risk_region} construct risk regions, while \cite{IPC} adopts a safe flight corridor that considers both static and dynamic obstacles. In \cite{raycast}, moving obstacle avoidance is achieved by Riemannian motion policy. Though some reactive methods enable reliable navigation, they become less effective in highly dynamic environments.

\textbf{Predictive methods:} Unlike reactive approaches, these methods utilize obstacle future predictions for planning. In \cite{ccmpc}\cite{vision-mpc}, a chance-constrained model predictive control approach is used to account for the uncertainties of dynamic obstacles. Wang et al. \cite{zju_collision_avoidance} enhance safe navigation by combining occlusion-aware obstacle tracking with spline-based trajectory optimization. Chen et al. \cite{risk-sampling} sample and evaluate trajectories based on predicted dynamic obstacles, while \cite{chen2023flying} incorporates UAV yaw angle for collision avoidance. In \cite{ViGO}, a vision-aided planner is used for avoiding both static and dynamic obstacles. For dynamic obstacle perception, a learning-based method \cite{eth-detector} is proposed for 3D detection and tracking. Additionally, the method proposed in \cite{frozone} uses learning-based approaches to avoid dynamic obstacles. However, it relies on a constant velocity model for predicting the future state of dynamic obstacles, which may not always be reliable. To better model pedestrian trajectories, Peddi et al. \cite{peddi2020data} trained a Hidden Markov Model on datasets, and Thomas et al. \cite{thomas2023foreseeable} applied self-supervised learning for map-level dynamic obstacle prediction. Although these methods, along with other popular learning-based approaches \cite{alahi2016social}\cite{gupta2018social}, improve upon traditional linear predictions, they primarily focus on low-level trajectory prediction and fail to account for the high-level decision-making processes of dynamic obstacles, potentially overlooking variations in obstacle behavior. Inspired by autonomous driving methods \cite{cui2019multimodal}\cite{katyal2020intent}\cite{benciolini2023non}\cite{zhou2023interaction} that predict multiple vehicle trajectories, we utilized the concept of dynamic obstacle intent in our planning framework. Our approach leverages obstacles' high-level decisions to forecast their trajectories and generate corresponding UAV trajectories for avoidance.
Unlike existing methods that aim for long-term predictions and require extensive sensing, our lightweight prediction is tailored to UAVs' limited sensory data and is integrated into our planning algorithm to enhance the safety of navigation.

\section{Problem Definition} \label{sec:prob_def}
This paper addresses the problem of UAV navigation in environments with dynamic obstacles, which primarily consist of human workers whose motion can be assumed to be non-holonomic. The UAV, equipped with an RGB-D camera, must navigate such environments by following a reference trajectory while avoiding collisions with both static and dynamic obstacles in $\mathbb{R}^{3}$ space. We define safe navigation as collision-free traversal and aim to minimize collisions. The UAV system adopts a linear dynamics model (Eqn. \ref{system_model}) where $I$ is a 3x3 identity matrix, \(\mathbf{p}, \mathbf{v}, \mathbf{a} \in \mathbb{R}^3\) represent position, velocity, and acceleration, respectively.

\begin{equation} \label{system_model}
\begin{bmatrix}
    \mathbf{p}_k \\
    \mathbf{v}_k
\end{bmatrix}
=
\begin{bmatrix}
    I & \Delta t I \\
    0 & I
\end{bmatrix}
\begin{bmatrix}
    \mathbf{p}_{k-1} \\
    \mathbf{v}_{k-1}
\end{bmatrix}
+ 
\begin{bmatrix}
    \frac{1}{2} \Delta t^2 I \\
    \Delta t I
\end{bmatrix}
\mathbf{a}_{k-1}
\end{equation}

\section{Methodology}
\begin{figure*}[t] 
    \centering
    \includegraphics[scale=0.644]{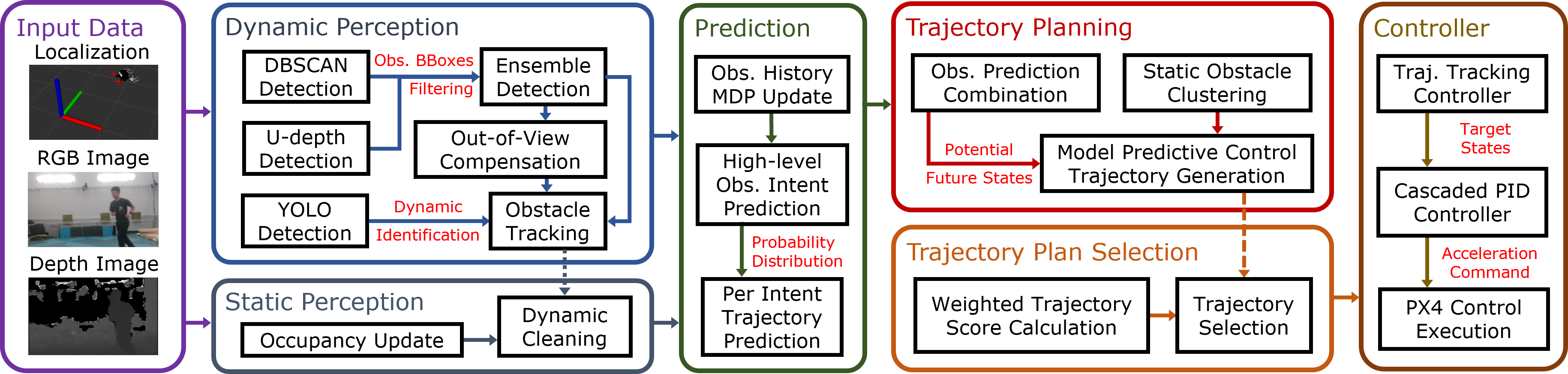}
    \caption{The proposed dynamic environment navigation system framework. Given localization, RGB, and depth image data, the static perception module constructs an occupancy map, while the dynamic perception module detects and tracks dynamic obstacles with out-of-view compensation. With the tracked obstacles' histories, the prediction module anticipates future obstacle trajectories using intent prediction. The intent-based planning algorithm,  including trajectory planning and selection, then optimizes and selects the highest-score trajectory. Finally, the selected trajectory is sent to the controller for execution. }
    \label{system-overview}
\end{figure*}
The proposed method contains three main modules: perception, prediction, and planning (Fig. \ref{system-overview}). The perception module (Sec. \ref{perception-section}) handles static and dynamic obstacles separately. The prediction module (Sec. \ref{prediction-section}) uses dynamic obstacle histories and a static occupancy map to generate intent probabilities and predicted trajectories. The intent-based planning algorithm (Sec. \ref{intent planning}) combines a model predictive control formulation (Sec. \ref{MPC-formulation}) with a score-based trajectory selection to produce the final execution trajectory.

\subsection{Lost-Tracking-Aware Perception} \label{perception-section}
The perception module contains static and dynamic components for obstacle perception. In Fig. \ref{system-overview}, the static perception uses a voxel map and dynamic cleaning to remove noise from moving obstacles. The dynamic perception, based on \cite{detector}, applies an updated ensemble detection method and out-of-view compensation to enhance accuracy and safety during collision avoidance, particularly in tracking loss scenarios.

\begin{figure}[t] 
    \centering
    \includegraphics[scale=0.78]{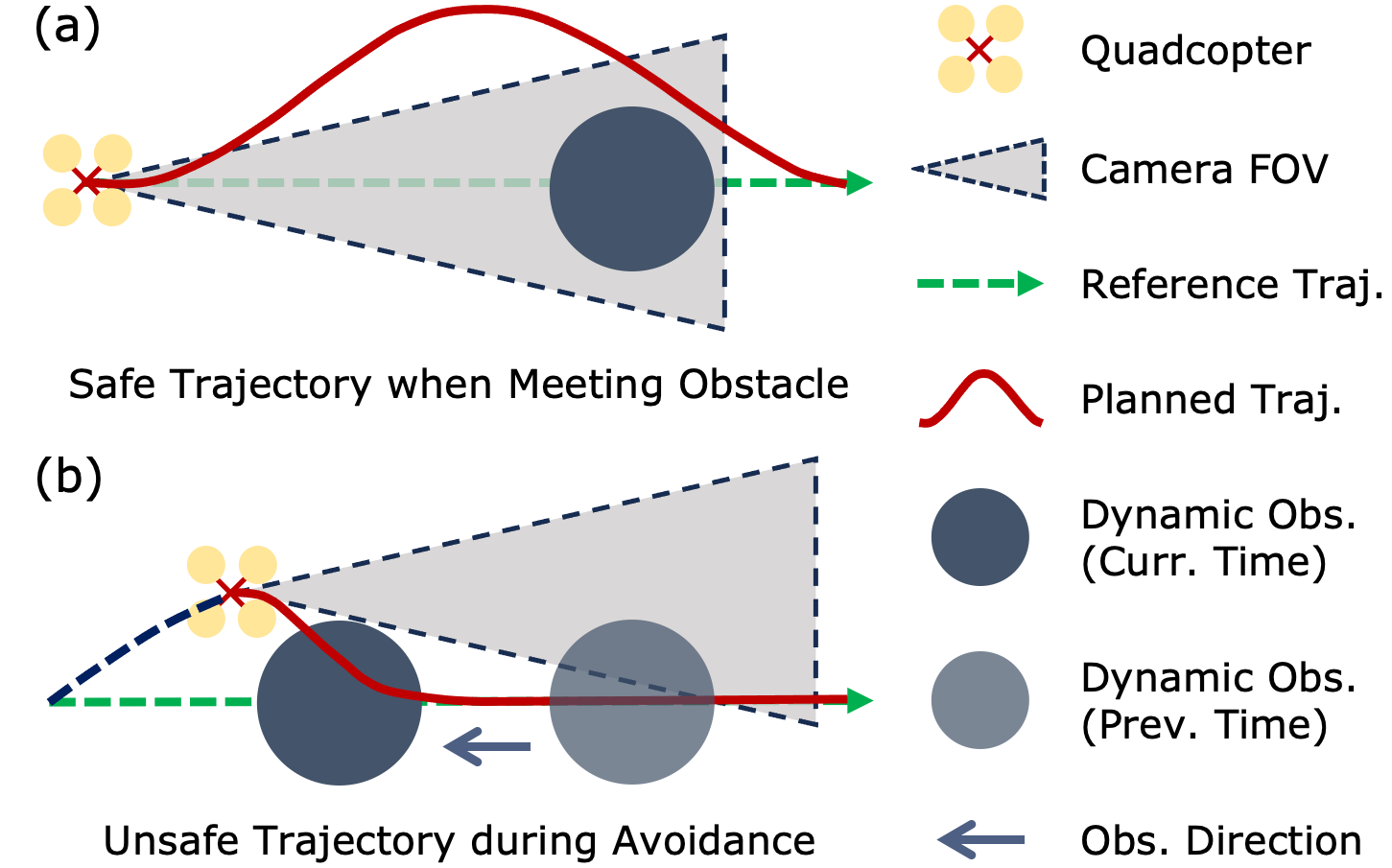}
    \caption{Illustration of unsafe dynamic obstacle avoidance due to tracking loss. (a) The robot initially detects the dynamic obstacle, generating a collision-free trajectory. (b) As the robot maneuvers, the obstacle moves out of the camera view, leading to an unsafe replanned trajectory.}
    \label{perception-illustration}
\end{figure}
\textbf{Dynamic Obstacle Detection: } 3D dynamic obstacle detection is challenging due to noisy depth images from UAV cameras and limited computational resources, making methods like \cite{pv-rcnn} impractical. To achieve high-accuracy results while maintaining the computational cost manageable, we propose an ensemble detection method that integrates two lightweight detectors. The first, known as the DBSCAN detector, uses the DBSCAN clustering algorithm \cite{DBSCAN} on point cloud data from depth images to determine obstacle centers and sizes based on the boundary points in each cluster. The second, the U-depth detector, processes raw depth images to create a U-depth map (similar to a top-down view) and uses a contiguous line-grouping algorithm \cite{early_reactive}\cite{vision-mpc} to detect 3D bounding boxes of obstacles. Note that, due to the noisy input data, both detectors can produce a significant number of false-positive results. The proposed ensemble detection method addresses this issue by identifying the mutual agreements between the two detectors. Because the detection results may include both static and dynamic obstacles, we first re-project the 3D bounding boxes (obtained from the mutual agreements between the DBSCAN and the U-depth detector) onto the 2D image plane. Next, we apply a lightweight YOLO detector to determine whether the re-projected 2D bounding boxes match the YOLO detections, using an IoU threshold for verification.  It is worth noting that, although the single YOLO detection approach on depth camera data can generate 3D detections, the resulting outputs may still be subject to errors stemming from noisy background data and misclassifications in the 2D detection process. Therefore, adopting an ensemble detection method is necessary for a robust perception system. We refer readers to \cite{detector} for detailed detection methods we build upon.

\textbf{Feature-based Tracking: } With the detected obstacles, feature-based tracking is used to estimate their states (i.e., positions and velocities). To minimize detection mismatches over time that can occur with the closest-center-point association method, a feature-based association method is proposed. The feature vector of an obstacle $\text{o}_{i}$ is defined as:
\begin{equation}
    f(\text{o}_{i}) = [\text{pos}(\text{o}_{i}),\  \text{dim}(\text{o}_{i}), \  \text{len}(\text{o}_{i}), \  \text{std}(\text{o}_{i})]^T,
\end{equation}
which includes information about the obstacle's position ($\text{pos}(\text{o}_{i})$), dimensions ($\text{dim}(\text{o}_{i})$), and the length ($\text{len}(\text{o}_{i})$) and standard deviation ($\text{std}(\text{o}_{i})$) of the obstacle's point cloud. To associate obstacles between the current and previous time steps, the two detected obstacles, $\text{o}_{i}$ and $\text{o}_{j}$,  with the highest similarity score are selected and calculated as:
\begin{equation}
    \text{sim}(\text{o}_{i}, \text{o}_{j}) = \text{exp}(-||W  (f(\text{o}_{i})-f(\text{o}_{j}))||_{2}^{2}),
\end{equation}
where $W$ is a diagonal matrix that weights the obstacle features. After the association process is complete, a linear Kalman filter with a constant acceleration model is applied to estimate the positions and velocities of the obstacles. Since the YOLO detectors can only classify dynamic objects included in their training dataset, the Kalman filter’s velocity estimation serves as an alternative check for identifying dynamic objects not present in the YOLO training set.

\textbf{Out-of-View Compensation: } Unlike stationary sensors, the onboard camera moves with the robot during collision avoidance. 
In Fig. \ref{perception-illustration}a, the robot may initially generate a collision-free trajectory with the obstacle within the camera's field of view. However, Fig. \ref{perception-illustration}b illustrates that the replanned trajectory becomes unsafe if the obstacle moves out of view during the maneuver. To address this issue, out-of-view compensation is introduced to estimate the states of obstacles when tracking is lost. Whenever a previously tracked dynamic obstacle leaves the sensor’s field of view, rather than discarding it entirely, its position is propagated using a constant-acceleration model that incorporates the estimated velocity and acceleration. In addition, its bounding size is gradually enlarged to account for increasing uncertainty over a short time horizon. These operations formulate a \say{risk region} that the planner should avoid for potential collision.

\subsection{Intent and Trajectory Prediction} \label{prediction-section}

\begin{figure}[t] 
    \centering
    \includegraphics[scale=0.292]{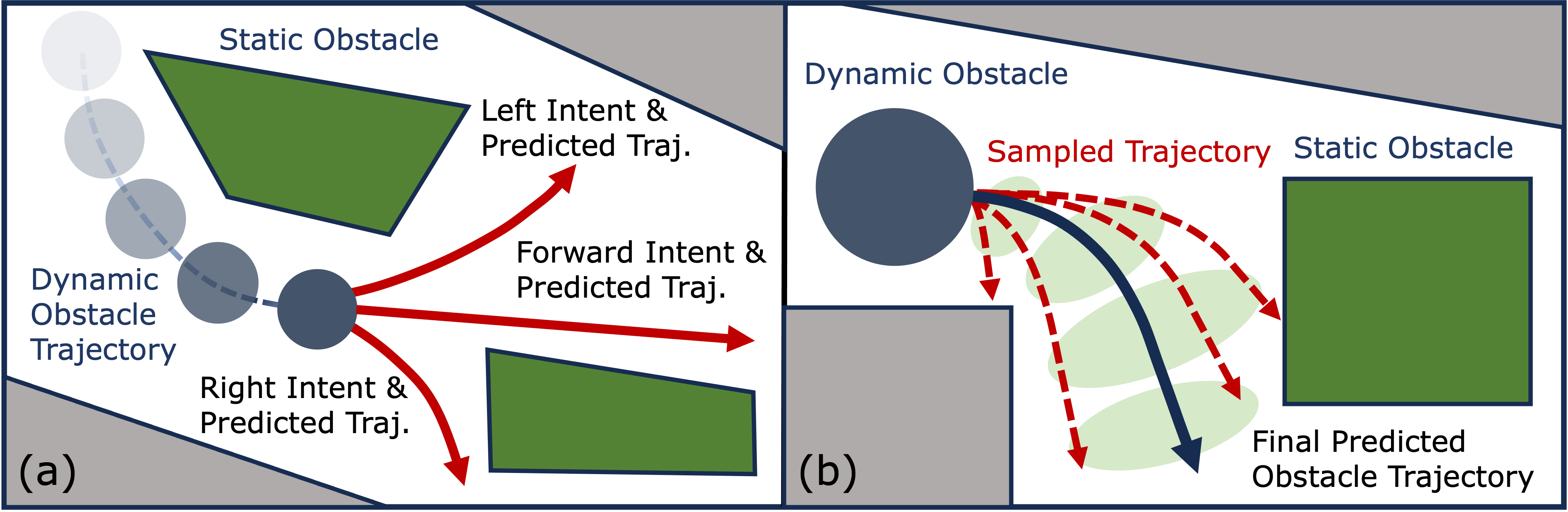}
    \caption{Illustration of the proposed prediction method. (a) Predicted trajectories (red curves) are generated from possible obstacle intents based on the obstacle's historical trajectory and the environment. (b) The example demonstrates trajectory prediction for the obstacle's right (direction) intent. The final predicted trajectories are represented by the mean of the sampled trajectories, with the trajectory variance indicated by shaded green areas. }
    \label{intent-figure}
\end{figure}

\textbf{Intent Prediction: } The various obstacle intents represent different high-level decisions of the dynamic obstacle. To model the changes in these decisions over time, we apply a Markov Decision Process (MDP), where the intents serve as the MDP states. The intent $\mathbb{I}^{\text{o}_{i}}_{t}$ of the $i$-th dynamic obstacle at time $t$ can be selected from a finite set $\mathcal{I}$ defined as:

\begin{equation}
\mathcal{I} = \{\text{forward}, \text{left}, \text{right}, \text{stop}\}, \ \mathbb{I}^{\text{o}_{i}}_{t} \in \mathcal{I}. 
\end{equation}
Here, we define four intents to consider the forwarding, turning, and stopping decisions of dynamic obstacles. Based on the historical trajectory of the $i$-th dynamic obstacle, our goal is to estimate the probability distribution $P^{o_{i}}_{\mathcal{T}}(\mathcal{I})$ over the set of possible intents at the current time $\mathcal{T}$. Note that dynamic obstacles are assumed to move in the 2D plane.

We begin by computing the raw intent probability $\Tilde{P}^{o_{i}}_{t}(\mathcal{I})$, which represents the intent probability distribution considering only the obstacle states at the time $t$ and the previous time step $t-1$. Given the obstacle states, the motion angle $\theta_{t}$ is defined as the angle between the displacement vectors of consecutive time steps, representing the change in facing direction between two consecutive time steps. The raw intent probability distribution $\Tilde{P}^{o_{i}}_{t}$ can be computed as: 
\begin{equation} \label{intent function}
\begin{split}
    \Tilde{P}^{o_{i}}_{t}(\mathcal{I}) = \frac{\hat{P}^{o_{i}}_{t}(\mathcal{I})}{\lVert \hat{P}^{o_{i}}_{t}(\mathcal{I}) \lVert}, \  \text{where} \  \hat{P}^{o_{i}}_{t}(\mathcal{I}) = 
    [e^{-\alpha \theta_{t}^2}, ... \\ \ \beta (1 + \sin{\theta_{t}}),
    \ \beta (1 - \sin{\theta_{t}}), \  1-\tanh{(\gamma \lVert v_{t} \rVert}) ]^T, 
\end{split}
\end{equation}
where the values of the raw intent probability distribution is calculated based on the order of the intent set $\mathcal{I}$, with $\alpha, \beta, \gamma$ being user-defined parameters that control the probability balance across various intents. The functions in Eqn. \ref{intent function} are heuristically chosen to model the probability of dynamic obstacle intents. For the stop intent, we use $1-\tanh(\cdot)$ to increase the stop probability as an obstacle's velocity approaches zero. The forward intent is modeled with an unnormalized Gaussian distribution based on motion angles which assigns higher probabilities to smaller angles that suggest straight-line movement. For the left and right intents, a sine-based function captures the increased likelihood of left or right turning with larger motion angles.

To fully leverage the trajectory of the dynamic obstacle, we use the raw intent probability $\Tilde{P}^{o_{i}}_{t}$ to first construct the unnormalized transition matrix $\hat{T}^{o_{i}}_{t}$ of the MDP at time $t$:
\begin{equation}
    \hat{T}^{o_{i}}_{t}  = \Sigma_{t} \cdot ( \Tilde{P}^{o_{i}}_{t}(\mathcal{I}) \otimes \textbf{1}^T), \ \ \Sigma_{t} \in \mathbb{R}^{4 \times 4}, 
\end{equation}
where $(\cdot  \otimes \textbf{1}^T )$ denotes an outer product operation in which each element of the vector is replicated across columns to form a square matrix. Meanwhile. $\Sigma_{t}$ is a diagonal matrix with all diagonal entries set to 1, except for the $n$-th entry.  Here, $n$ indicates the intent index with the highest probability at the previous time step. This $n$-th entry is assigned a user-defined weighting parameter $s$ (with $s>1$), thereby increasing the likelihood of the obstacle continuing its previously dominant intent. The transition matrix $T^{o_{i}}_{t}$ is then obtained by normalizing each row of $\hat{T}^{o_{i}}_{t}$.  

Finally, by initializing the intent probability $P^{o_{i}}_{0}$ at $t=0$ to a uniform distribution, we can apply the MDP to obtain the intent probability distribution $P^{o_{i}}_{\mathcal{T}}$ at current time $\mathcal{T}$ using: 
\begin{equation}
    P^{o_{i}}_{\mathcal{T}} = (\prod_{t=0}^{\mathcal{T}} T^{o_{i}}_{t}) \cdot P^{o_{i}}_{0}, \ \text{where} \  \mathcal{T} = n\Delta t.
\end{equation}
This process is computed at each time step with only a limited duration of each dynamic obstacle's history retained. The computed intent probability distribution is utilized by the planning algorithm, which will be discussed in later sections.

\textbf{Trajectory Prediction: } The trajectory-level prediction computes the future positions and risk sizes of the obstacle for all possible intents, as outlined in Alg. \ref{trajectory prediction}. It is important to note that for each obstacle, trajectories are predicted for all four intents, regardless of the intent probability distribution. For the stop intent (Lines \ref{stop start}-\ref{stop end}), the predicted positions replicate the current obstacle position, while the risk sizes are inflated based on the obstacle's velocity and the time elapsed. For the forward, left, and right intents, a set of acceleration-based control inputs is generated (Lines \ref{control start}-\ref{control end}). Constant linear and angular acceleration are then applied to propagate the obstacle positions. The forward intent generates controls by varying the linear acceleration within a predefined range (Line \ref{forward control}, whereas the left and right intents adjust both angular and linear acceleration within their respective predefined ranges (Line \ref{turning control}).

By applying these control combinations to the obstacle (Lines \ref{propagate control start}-\ref{propagate control end}) with propagation continuing until a collision occurs, multiple trajectories are sampled for each intent. Since each obstacle’s multiple intents are evaluated separately, we assume trajectories associated with a single intent remain closely clustered. Moreover, the short prediction intervals ensure these trajectories remain similar. Consequently, the predicted trajectory is determined by averaging the positions of all sampled trajectories at each time step, and the risk sizes of the obstacle are inflated by adding a value proportional to the standard deviation of the sampled trajectory positions (Lines \ref{output pos}-\ref{output size}). If the mean trajectory is not collision-free, a sampled trajectory closest to the mean trajectory will be selected. Fig. \ref{intent-figure}b visualizes the sampled trajectories for the right intent as red dotted curves and the final predicted trajectory as a blue solid curve.

\begin{algorithm}[t] \label{trajectory prediction}
\caption{Trajectory-level Prediction} 
\SetAlgoNoLine%
$o \gets \text{Input Dynamic Obstacle}$\;
$N_{\text{pred}} \gets \text{Prediction Time Steps}$\;
$p^{o}, v^{o}, a^{o}, s^{o} \gets \textbf{obsInfo}(o)$ \Comment*[r]{pos, vel, accel, size}
$\text{predictions} \gets \emptyset$ \Comment*[r]{all intent predictions}
\For{$\mathbb{I}^{o}$ \normalfont{\textbf{in}} $\mathcal{I}$}{ 
    $\mathcal{P}_{\text{pos}}, \mathcal{P}_{\text{size}} \gets  \emptyset $ \Comment*[r]{prediction array}
    \If{$\mathbb{I}^{o}$ \normalfont{\textbf{is}}  \text{stop}}{ \label{stop start}
        \For{$\text{n}$ \normalfont{\textbf{in}} $\normalfont{\textbf{range}}(0, N_{\text{pred}})$}{
            $\mathcal{P}_{\text{pos}}.\normalfont{\textbf{append}}(p^{o})$\;
            $\mathcal{P}_{\text{size}}.\normalfont{\textbf{append}}(s^{o} + n\Delta t \cdot \normalfont{\textbf{min}}(v^{o}, v_{\text{thresh}}))$\; \label{stop end}
        }
        $\normalfont{\text{predictions}}.\normalfont{\textbf{append}}(\{\mathcal{P}_{\text{pos}}, \mathcal{P}_{\text{size}}\})$\;
        \normalfont{\textbf{continue}}\;
    }
    \If{$\mathbb{I}^{o}$ \normalfont{\textbf{is}}  \text{forward}}{ \label{control start}
        $\mathcal{S}_{\text{control}} \gets \normalfont{\textbf{getLinearControlSet}}(a^{o}) $;\ \label{forward control}
    }
    \ElseIf{$\mathbb{I}^{o}$ \normalfont{\textbf{is}}  \text{left} \normalfont{\textbf{or}}  $\mathbb{I}^{o}$ \normalfont{\textbf{is}}  \text{right}}{ 
        $\mathcal{S}_{\text{control}} \gets \normalfont{\textbf{getTurningControlSet}}(a^{o}, \mathbb{I}^{o}) $;\ \label{turning control} \label{control end}
    }
    $\mathcal{S}_{\text{traj}} \gets \emptyset$ \Comment*[r]{pred trajectory array} 
    \For{$\text{c}_{i}$ \normalfont{\textbf{in}} $\mathcal{S}_{\text{control}}$}{ \label{propagate control start}
        $\sigma_{\text{traj}} \gets \normalfont{\textbf{getConstControlTraj}(p^{o}, c_{i}, N_{\text{pred}}})$\;
        $\mathcal{S}_{\text{traj}}.\normalfont{\textbf{append}}(\sigma_{\text{traj}}) $\; \label{propagate control end} 
    }
    $\mathcal{P}_{\text{pos}}, \mathcal{V}_{\text{pos}} \gets \normalfont{\textbf{getPositionMeanAndVar}}(\mathcal{S}_{\text{traj}})$\; \label{output pos}
    $\mathcal{P}_{\text{size}} \gets \lambda \cdot \sqrt{\mathcal{V}_{\text{pos}}} + s^{o}$\;
    $\normalfont{\text{predictions}}.\normalfont{\textbf{append}}(\{\mathcal{P}_{\text{pos}}, \mathcal{P}_{\text{size}}\})$\; \label{output size}
}
\textbf{return} \text{predictions}\;
\end{algorithm}

\subsection{MPC Formulation for Trajectory Generation} \label{MPC-formulation}
Given a reference trajectory, we apply the model predictive control method to generate local trajectories to avoid collisions. The entire optimization problem can be formulated as:
\begin{mini!}[2]
    {\mathbf{x}_{0:N}, \mathbf{u}_{0:N-1}}{\sum_{k=0}^{N} {\begin{Vmatrix} \mathbf{x}_{k} - \mathbf{x}_{k}^\text{ref} \end{Vmatrix}^2 + \sum_{k=0}^{N-1} \lambda_{\mathbf{u}}\begin{Vmatrix} \mathbf{u}_{k} \end{Vmatrix}^2},}{}{} \label{mpc objective}
\addConstraint{\mathbf{x}_{0}}{=\mathbf{x}(t_{0})}{} \label{initial constraint}
\addConstraint{\mathbf{x}_{k}}{=f(\mathbf{x}_{k-1}, \mathbf{u}_{k-1})}{} \label{dynamics model}
\addConstraint{\mathbf{u}_{\text{min}} \leq}{\mathbf{u}_{k} \leq  \mathbf{u}_{\text{max}}}{} \label{control limits}
\addConstraint{\mathbf{x}_{k} \not\in C_{i}}{, \forall i \in S^{\text{static}}_{o} \cup S^{\text{dynamic}}_{o}}{} \label{collision constraint}
\addConstraint{\forall k \in \{0, \ldots , N\}}, 
\end{mini!}
where $\mathbf{x_k} = [\mathbf{p_k},\mathbf{v_k}]^T$ and $\mathbf{u_k} = \mathbf{a_k}$ represent the robot states and control inputs with the subscript indicating the time step. The detailed dynamic model is provided in Sec. \ref{sec:prob_def}. The objective (Eqn. \ref{mpc objective}) is to minimize deviation from the reference trajectory while using the least control effort.  Eqn. \ref{initial constraint} sets the initial state constraint based on the current robot states. The robot's dynamics model and control limits are presented by Eqns. \ref{dynamics model} and \ref{control limits}, respectively.

The collision constraints (Eqn. \ref{collision constraint}) ensure that the robot avoids collisions with static and dynamic obstacles. 
Static obstacles are initially represented in an occupancy voxel map, while dynamic obstacles use axis-aligned 3D bounding boxes. To unify these representations, we convert static obstacles into oriented 3D bounding boxes using a hierarchical clustering method. First, we convert the occupancy voxels into a point cloud and then apply DBSCAN clustering to determine bounding box centroids and dimensions for obstacle clusters. These bounding boxes are then refined by iteratively applying K-means++, starting with two centroids at the box corners, until the point-cloud density within each box exceeds a specified threshold. Finally, the 2D orientation angle $\phi$ of each static bounding box is discretely adjusted to maximize point-cloud density. With both static and dynamic obstacles uniformly represented as 3D bounding boxes, the collision constraint for each obstacle can be expressed as:
\begin{equation}
    \frac{(x \cos{\phi} + y \sin{\phi})^2}{a^2} + 
    \frac{(-x \sin{\phi} + y \cos{\phi})^2}{b^2} + 
    \frac{z^2}{c^2} \geq 1,
\end{equation}
where $a$, $b$, and $c$ are the half-lengths of the principal axes of the minimum ellipsoid enclosing the obstacle's risk-size bounding box, and $x$, $y$, and $z$ are the relative positions along the three axes between the robot and the obstacle centroid. 

\subsection{Intent-Based Trajectory Planning} \label{intent planning}
\begin{algorithm}[t] \label{trajectory planning}
\caption{Intent-based Trajectory Planning} 
\SetAlgoNoLine%
$\mathcal{M} \gets \text{Static Occupancy Map}$\;
$p^{r} \gets \text{Current Robot Position}$\;
$\mathbf{\sigma}_{\text{ref}} \gets \text{Input Reference Trajectory}$\;
$\mathbf{N}_{ic} \gets \text{Number of Intent Combinations}$\;
$\mathcal{S}_{so} \gets \normalfont{\textbf{clusterStaticObstacles}}(\mathcal{M}, p^{r})$\; \label{get static}
$\mathcal{S}_{do} \gets \normalfont{\textbf{closeDynamicObstacles}}(p^{r})$\; \label{get dynamic}
$\mathcal{S}_{ic} \gets \normalfont{\textbf{topIntentCombinations}}(\mathcal{S}_{do}, N_{ic})$\;
$\sigma_{\text{output}} \gets \emptyset, \ \text{highestScore} \gets 0$\;
\For{$n$ \normalfont{\textbf{in}} $\normalfont{\textbf{range}}(0, N_{ic})$}{ 
    $\mathcal{IC} \gets  \mathcal{S}_{ic}[n] $ \Comment*[r]{Intent Combination} \label{intent combination}
    $\mathcal{S}^{\text{pred}}_{\text{do}} \gets \normalfont{\textbf{getObstaclePredictions}}(IC, \mathcal{S}_{do})$\; \label{traj pred}
    $\sigma_{\text{traj}} \gets \normalfont{\textbf{MPCTrajGeneration}}(\sigma_{\text{ref}}, \mathcal{S}_{so} , \mathcal{S}^{\text{pred}}_{do} )$\; \label{mpc traj}
    $\text{score} \gets P_{ic} \cdot \normalfont{\textbf{evaluateTraj}}(\sigma_{\text{traj}})$\; \label{score begin}
    \If{$\normalfont{\text{score}} > \normalfont{\text{highestScore}}$}{
        $\sigma_{\text{output}} \gets \sigma_{\text{traj}} $\;
        $\text{highestScore} \gets \text{score}$\; \label{score end}
    }
}
$\textbf{return}\  \sigma_{\text{output}}$\;
\end{algorithm}

The proposed algorithm (Alg. \ref{trajectory planning}) generates multiple trajectories corresponding to different obstacle intent combinations and selects the optimal one based on an evaluation system. Initially, static obstacles are clustered into a set of bounding boxes $\mathcal{S}_{so}$, and the set of risky dynamic obstacles $\mathcal{S}_{do}$ near the robot is identified (Lines \ref{get static}-\ref{get dynamic}). Next, the top $N_{ic}$ intent combinations $S_{ic}$ for different obstacles are determined based on their current intent probability distributions. Each intent combination (labeled as $\mathcal{IC}$ in Line \ref{intent combination}) represents a possible future action scenario for all obstacles, and the algorithm focuses on the most likely $N_{ic}$ scenarios. The likelihood of each combination is computed by multiplying the individual intent probabilities of each obstacle, assuming the distributions are independent across obstacles. For a single intent combination, the trajectory-level prediction is obtained for each obstacle (Line \ref{traj pred}), and MPC is used to generate a collision-free trajectory $\sigma_{\text{safeTraj}}$ for navigation (Line \ref{mpc traj}). The score of the generated trajectory is calculated by combining the intent combination probability $P_{ic}$ with the raw score from the evaluation system (Lines \ref{score begin}-\ref{score end}). The algorithm selects the highest-score trajectory for execution.

The evaluation system computes the raw trajectory score using a weighted sum of three components: the consistency score, the detouring score, and the safety score:
\begin{equation}
    \mathbb{S}_{\text{traj}} = \lambda_{\text{1}} \cdot \mathbb{S}_\text{cons} + \lambda_{\text{2}} \cdot \mathbb{S}_\text{detour} + \lambda_{\text{3}} \cdot \mathbb{S}_\text{safety}, 
\end{equation}
where $\lambda_{1}$, $\lambda_{2}$, and $\lambda_{3}$ are user-defined weights. The consistency score evaluates how closely the new trajectory matches the previous one to prevent oscillating behaviors:
\begin{equation}
    \mathbb{S}_\text{cons} = \frac{N}{\sum_{k=1}^{N} \lVert \sigma_{\text{traj}}(k) - \sigma_{\text{traj}}^{\text{prev}}(k)\rVert}.
\end{equation}
Similarly, the detouring score is calculated based on the point-wise distance between the trajectory and the reference:
\begin{equation}
    \mathbb{S}_\text{detour} = \frac{N}{\sum_{k=1}^{N} \lVert \sigma_{\text{traj}}(k) - \sigma_{\text{ref}}(k) \rVert} .
\end{equation}
Both the consistency score and the detouring score are capped at a maximum allowable value to prevent extremely high scores when the denominators approach zero. However, due to the motion uncertainties of dynamic obstacles, predictions cannot always be guaranteed to be valid. To further reduce the risk of collisions while preserving consistency with previous trajectories, the trajectory is preferred when it maintains a sufficient distance from obstacles. The safety score is therefore determined by the distance to static and dynamic obstacles under the most likely intent combination:
\begin{equation}
    \mathbb{S}_\text{safety} = \frac{1}{N} \sum_{k=1}^{N} \frac{1}{N_{so} + N_{do}} \sum_{i=1}^{N_{so} + N_{do}} \lVert \sigma_{\text{traj}}(k) - p^{o_{i}}_{k}  \rVert.
\end{equation}
These scores work together to ensure the trajectory maintains smoothness, follows the reference, and maximizes safety.

\section{Result and Discussion}
To evaluate the proposed method, we conduct simulation and flight tests in dynamic environments. An Intel RealSense D435i camera is used for both static and dynamic perception, and the LiDAR Inertial Odometry (LIO) \cite{xu2022fast} is adopted for state estimation. Physical flight tests run on an NVIDIA Orin NX onboard computer. Experiments use a maximum velocity of 1.5m/s per axis, a 0.1m resolution occupancy map, and a 3.0s prediction time for dynamic obstacles.
A time step of 0.1s and a horizon of 30 are used for MPC.

\subsection{Simulation Experiments}
\begin{figure}[t] 
    \centering
    \includegraphics[scale=0.76]{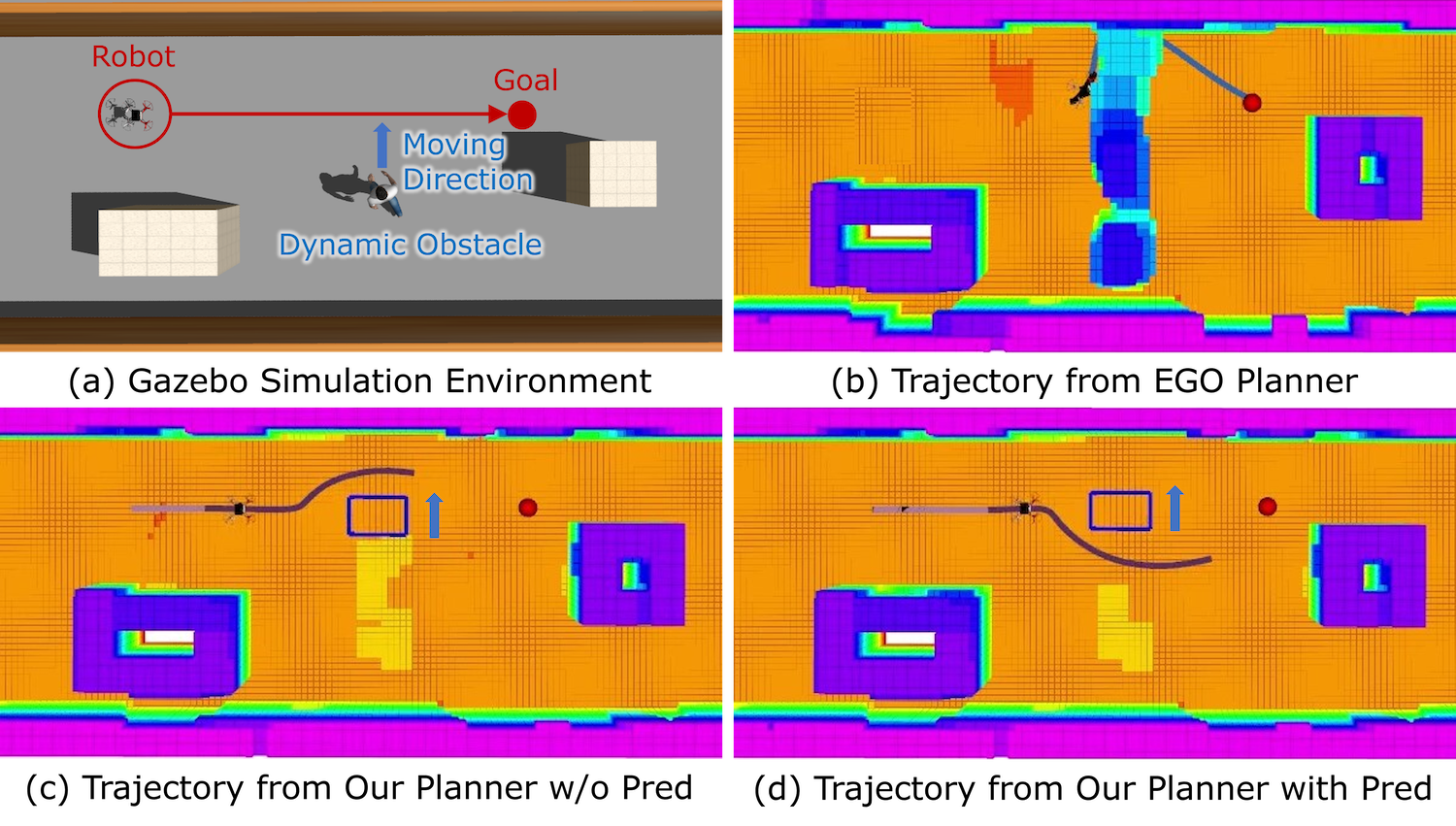}
    \caption{Comparison of generated trajectories from benchmark planners in the same scenario. (a) Experiment scenario in Gazebo. (b) EGO Planner \cite{ego_planner} failure due to a noisy map. (c) Our planner's trajectory without prediction, resulting in potential collision. (d) Safe trajectory generated by our planner.}
    \label{sim-experiments}
\end{figure}

The qualitative experiments, in which dynamic obstacles follow predefined polynomial trajectories to simulate human motion with varying velocity and intent, demonstrate the need to integrate dynamic perception and prediction while highlighting the prediction module’s effectiveness. 
For example, Fig. \ref{sim-experiments}a illustrates a scenario where the robot navigates to its goal while a dynamic obstacle crosses its path. Using the EGO planner \cite{ego_planner}, trajectory generation fails due to noisy occupancy data (Fig. \ref{sim-experiments}b). In contrast, our planner successfully generates a safe trajectory with a cleaned map and tracked dynamic obstacle bounding box (Fig. \ref{sim-experiments}d), showing that static occupancy updates are insufficient and emphasizing our combined perception module. When comparing the trajectories generated with and without the prediction module, Fig. \ref{sim-experiments}c clearly shows that the trajectory without prediction leads to a potential future collision with the dynamic obstacle moving in the direction of the blue arrow, while the prediction module enables the generation of a trajectory that safely avoids future collisions.

\begin{figure}[t] 
    \centering
    \includegraphics[scale=0.293]{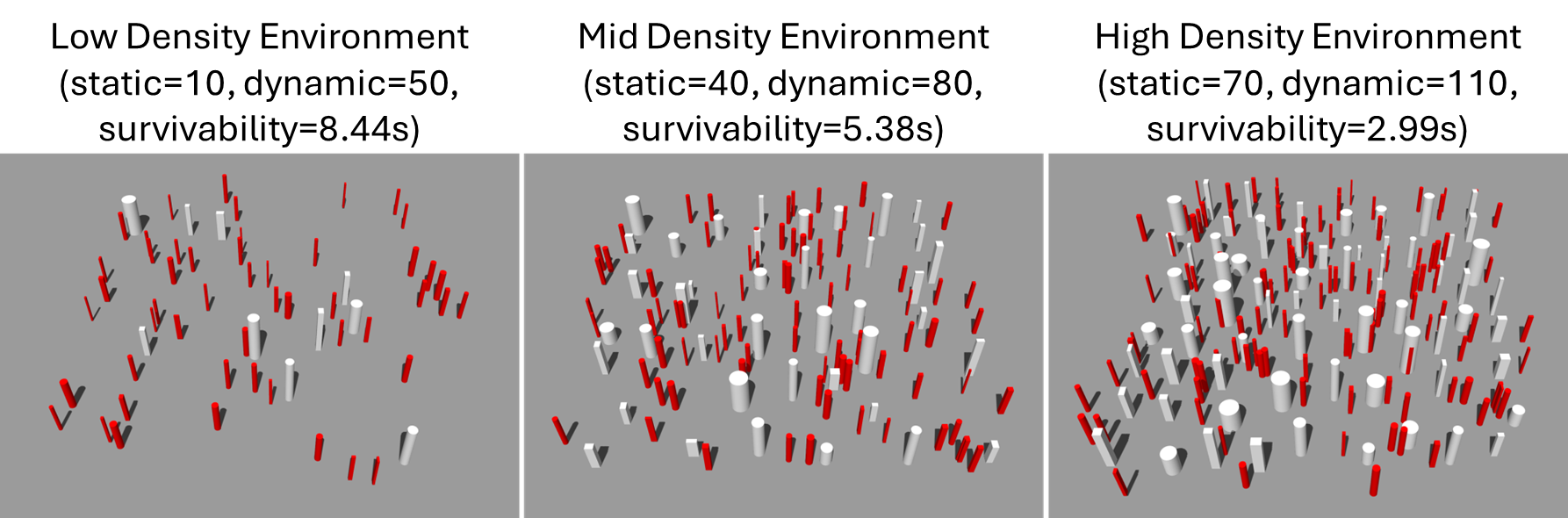}
    \caption{Experiment environments with varying obstacle densities. Static obstacles are shown in white, and dynamic obstacles are displayed in red.}
    \label{collision-experiment-env}
\end{figure}

The quantitative results show the number of collisions for our system and benchmark methods across environments with different difficulty levels (Fig. \ref{collision-experiment-env}). The difficulty metric, survivability, are calculated as described in \cite{dynamic_metircs}. The white cylinders represent static obstacles, while the red cylinders denote dynamic obstacles moving at a constant velocity toward shifting goals, resulting in non-linear trajectories. We benchmark against the EGO planner \cite{ego_planner}, ViGO \cite{ViGO}, and CC-MPC \cite{vision-mpc} and evaluate our method with and without the prediction module and safety score. Each method was tested 20 times per environment, with collision counts shown in Table \ref{collision rate result} (percentages relative to ViGO in parentheses). Overall, our planner demonstrates the fewest collisions across all three environments. The N/A value for the EGO planner indicates that its trajectory generation fails due to noisy maps in the medium and high-density environments. Even without the prediction module, our approach outperforms ViGO, whose B-spline optimization lacks agility for close-proximity obstacles, whereas our MPC-based method enables more responsive avoidance. CC-MPC maintains a relatively low collision rate but experiences solver failures in high-density environments due to its strict chance constraints. Instead of relying on such strict constraints, our method improves safety by planning with multiple intents, leading to fewer solver failures in high-density environments, thus fewer collisions. The comparison of our method with and without the prediction module shows that prediction effectively reduces collisions, aligning with qualitative observations. Similarly, removing the safety score results in slightly higher collision rates in low and medium-density environments but a significant increase in high-density scenarios, highlighting its role in trajectory selection. Additionally, comparing our method without the prediction module to the version with prediction (but without the safety score) confirms that prediction significantly lowers collision rates across all environments.

\begin{table}[t]
\begin{center}
\caption{Benchmark of the number of collisions from 20 sample runs in environments with different obstacle densities.} \label{collision rate result}
\begin{tabular}{ |l | c | c| c |  } 
 \hline

 \multicolumn{4}{|c|}{Collision Times Measurement with Benchmarks on a 20m$\times$20m Map} \Tstrut\\
 \hline

 \multirow{2}{*}{Obstacle Density}  &  Low Density &  Mid Density  & High Density \Tstrut\\ 
  & S=10, D=50 & S=40, D=80 & S=70, D=110 \\
 \hline

 EGO Planner \cite{ego_planner}  & 53 ($196.2\%$)  & N/A & N/A \Tstrut\\ 
 \hline
 ViGO \cite{ViGO} & 27 ($100\%$) & 56 ($100\%$) &  69 ($100\%$) \Tstrut\\  
 \hline
 CC-MPC \cite{vision-mpc} & 16 ($59.3
 \%$) & 32 ($57.1\%$) & 73 ($105.7\%$)  \Tstrut\\ 
 \hline
 Ours w/o pred & 12 ($44.4\%$) & 25 ($44.6\%$) & 43 ($62.3\%$) \Tstrut\\  
 \hline
 Ours w/o safety & 5 ($18.5
 \%$) & 12 ($21.4\%$) & 20 ($30.0\%$)  \Tstrut\\ 
 \hline
 \textbf{Ours}  & \textbf{4} ($\textbf{14.8\%}$) & \textbf{11} ($\textbf{19.6\%}$) & \textbf{14} ($\textbf{20.3\%}$) \Tstrut\\ 
 \hline
\end{tabular}
\end{center}
\end{table}

\subsection{Physical Flight Tests}
\begin{figure}[t] 
    \centering
    \includegraphics[scale=1.15]{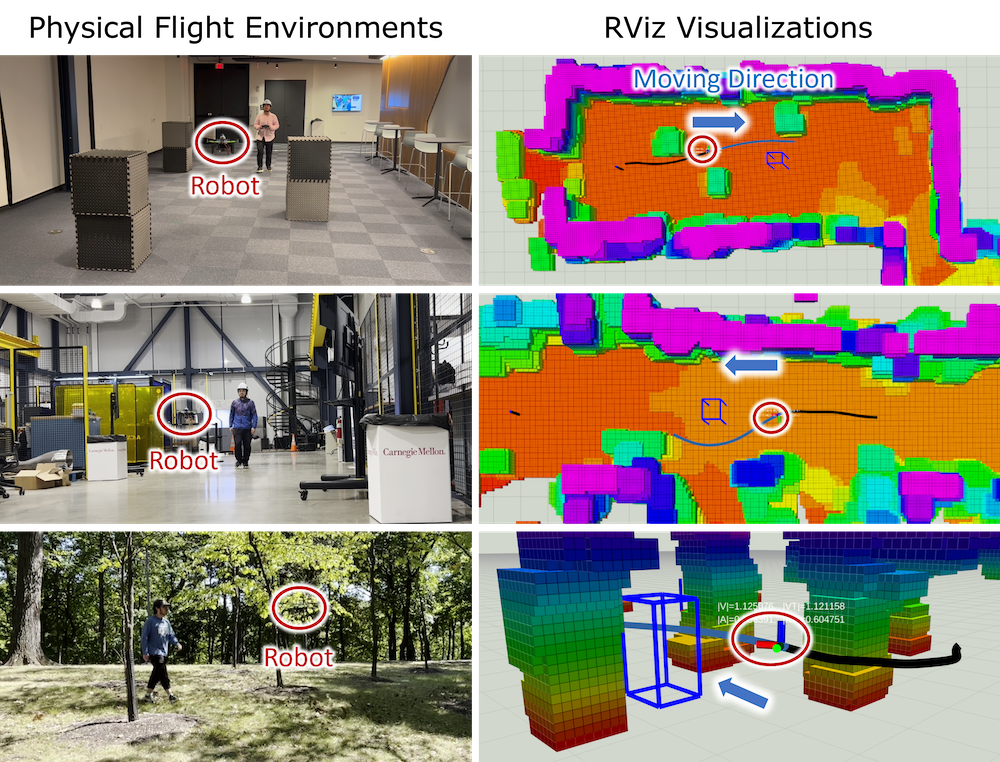}
    \caption{Physical flight experiments. The left column figures show moments of collision avoidance in various environments, while the right column figures visualize the robot following safe trajectories to avoid obstacles.}
    \label{flight-experiments}
\end{figure}

To validate the system’s real-world performance, we conducted physical flight tests in two indoor and one outdoor environments, as shown in Fig.\ref{flight-experiments}. In each environment, the robot either builds the map in real-time for straight-line navigation or uses a pre-built map with waypoints to navigate through complex structures. To simulate dynamic obstacles, we let persons intentionally walk toward the robot at a normal walking speed, blocking the robot's paths.  The left column of Fig. \ref{flight-experiments} shows the robot encountering humans, while the right column displays the safe trajectories generated for collision avoidance. Our observations confirm that the robot navigates safely in all environments, demonstrating the effectiveness of the proposed method.

Since the proposed dynamic perception module uses a camera with a limited field of view, we also performed experiments using a motion capture system to verify obstacle avoidance ability from all directions (see supplementary video). In the experiment, the robot hovers in place while multiple pedestrians approach, successfully avoiding them and returning to its original position once they move away.

During experiments, we measured the computation time for each module of the proposed navigation system. All modules run in parallel on separate threads. The perception module takes 15ms for static and 27ms for dynamic perception. The dynamic obstacle prediction module requires 0.05ms for intent prediction and 2ms for trajectory prediction. The planning module computes a single MPC trajectory in 10ms. The computational time for intent-based planning, which includes multiple MPC trajectories, depends on the number of obstacles but remains manageable due to the limited number of local obstacles. On average, intent-based planning takes 68ms. These results demonstrate that the system operates within real-time constraints, enabling effective navigation and obstacle avoidance in dynamic environments.

\subsection{Intent-based Trajectory Prediction Evaluation}
We evaluated our intent-based trajectory prediction using motion capture experiments to obtain ground truth trajectories and measure average and final displacement errors (ADE and FDE) (Table \ref{prediction experiment}).
We let persons walk in front of the robot's camera to simulate collision avoidance scenarios. The experiment scenarios were divided into two categories: with static obstacles and without static obstacles. For each scenario, over 100 dynamic obstacle trajectories were collected and compared between linear prediction and our intent-based method, utilizing both ground truth and perception module data. Our intent-based method generates multiple trajectories based on different intents, so we calculated displacement errors by selecting the trajectory closest to the ground truth. This evaluation approach is reasonable because our planning algorithm considers all predicted trajectories when generating plans. Therefore, if any of the predicted trajectories is accurate, the planner can use it to ensure effective navigation. Table \ref{prediction experiment} shows that our intent-based prediction method outperforms linear prediction by achieving lower ADE and FDE which enables more effective collision avoidance. Additionally, the prediction has a slightly higher error when using the perception module compared to the ground truth obstacle data. Furthermore, environments without static obstacles present higher prediction errors due to the less constrained motion of dynamic obstacles, which increases the difficulty of accurate trajectory prediction. 

\begin{table}[t]
\begin{center}
\caption{Evaluation of average and final displacement errors for linear and intent-based trajectory prediction methods.} \label{prediction experiment}
 \begin{tabular}{ |l | c | c| c | c | } 
 \hline
 Experiments & \multicolumn{2}{|c|}{w/o static obstacles} & \multicolumn{2}{|c|}{w/ static obstacles} \Tstrut\\
 \hline
 Metrics [meter]  & ADE  & FDE & ADE & FDE  \Tstrut\\ 
 \hline
 Linear w/ mocap & 1.339 & 3.211 & 0.840  & 1.965 \Tstrut\\  
 \hline
 Ours w/ mocap & 0.699 & 1.264 & 0.543 & 1.064 \Tstrut\\  
 \hline
 Linear w/ perception & 1.554 & 3.611 & 0.899 &  2.116 \Tstrut\\ 
 \hline
 Ours w/ perception & 0.805 & 1.397 & 0.552 & 1.104 \Tstrut\\ 
 \hline
\end{tabular}
\end{center}
\end{table}

\section{Conclusion and Future Work}
This paper introduces an autonomous navigation framework that enhances the safety and effectiveness of UAV navigation in dynamic environments. 
Our approach addresses key challenges in perception and planning by integrating a dynamic obstacle intent prediction mechanism. Specifically, we use a perception module to efficiently detect and track dynamic obstacles and an intent-based prediction module to forecast their actions and trajectories. To reduce collisions, we propose an intent-based planning algorithm that leverages model predictive control (MPC) to compute optimal trajectories. Simulation experiments and physical flight tests confirm that the proposed framework improves the safety of navigation in dynamic environments. Our future work will focus on integrating a LiDAR to extend the perception range.

\bibliographystyle{IEEEtran}
\bibliography{IEEEabrv,bibliography.bib}

\end{document}